\documentclass[10pt,twocolumn,letterpaper]{article}
\usepackage[accsupp]{axessibility}
\usepackage{cvpr}              

\usepackage[dvipsnames]{xcolor}

\usepackage{algorithm,algpseudocode}
\usepackage{command}

\definecolor{cvprblue}{rgb}{0.21,0.49,0.74}
\usepackage[pagebackref,breaklinks,colorlinks,citecolor=cvprblue]{hyperref}

\def\paperID{4795} 
\def\confName{CVPR}
\def\confYear{2024}

\title{Boosting Adversarial Transferability by Block Shuffle and Rotation}

\author{
Kunyu Wang\textsuperscript{\rm 1}, Xuanran He\textsuperscript{\rm 2}, Wenxuan Wang\textsuperscript{\rm 1}, Xiaosen Wang\textsuperscript{\rm 3}\thanks{Corresponding author. Email: \texttt{xiaosen@hust.edu.cn}}\\
\textsuperscript{\rm 1}Chinese University of Hong Kong, 
\textsuperscript{\rm 2}Nanyang Technological University, \\
\textsuperscript{\rm 3}Huawei Singularity Security Lab \\
}

\begin{document}
\maketitle
\begin{abstract}
   Adversarial examples mislead deep neural networks with imperceptible perturbations and have brought significant threats to deep learning. An important aspect is their transferability, which refers to their ability to deceive other models, thus enabling attacks in the black-box setting. Though various methods have been proposed to boost transferability, the performance still falls short compared with white-box attacks. In this work, we observe that existing input transformation based attacks, one of the mainstream transfer-based attacks, result in different attention heatmaps on various models, which might limit the transferability. We also find that breaking the intrinsic relation of the image can disrupt the attention heatmap of the original image. Based on this finding, we propose a novel input transformation based attack called block shuffle and rotation (\name). Specifically, \name splits the input image into several blocks, then randomly shuffles and rotates these blocks to construct a set of new images for gradient calculation. Empirical evaluations on the ImageNet dataset demonstrate that \name could achieve significantly better transferability than the existing input transformation based methods under single-model and ensemble-model settings. Combining \name with the current input transformation method can further improve the transferability, which significantly outperforms the state-of-the-art methods. Code is available at \url{https://github.com/Trustworthy-AI-Group/BSR}.
\end{abstract}
\section{Introduction}
\label{sec:intro}
Deep neural networks (DNNs) have established superior performance in many tasks, such as image classification~\cite{he2016resnet,huang2017densely}, segmentation~\cite{jonathan2015fully}, object detection~\cite{ren2015faster,redmon2016you}, face recognition~\cite{wang2018cosface}, among others. Despite their achievements, DNNs have been observed to exhibit significant vulnerability to adversarial examples~\cite{szegedy2014intriguing,goodfellow2015FGSM,wang2019atgan}, which closely resemble legitimate examples, yet can deliberately misguide deep learning models to produce unreasonable predictions. The existence of such vulnerabilities gives rise to serious concerns, particularly in security-sensitive applications, such as autonomous driving~\cite{eykholt2018robust}, face verification~\cite{sharif2016accessorize}.

\begin{figure}
    \centering
    \begin{minipage}[c]{0.15\textwidth} 
          \centering 
          \includegraphics[width=\linewidth]{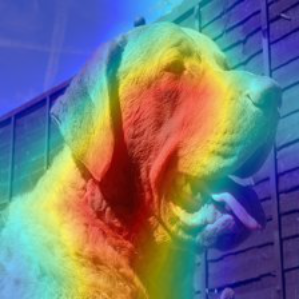}
          \caption*{\small Raw Image}
    \end{minipage}
    \hspace{-0.2em}
    \begin{minipage}[c]{0.15\textwidth} 
          \centering 
          \includegraphics[width=\linewidth]{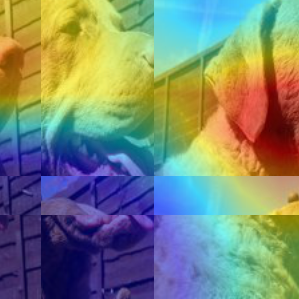}
          \caption*{\small Shuffled Image}
    \end{minipage}
    \hspace{-0.2em}
    \begin{minipage}[c]{0.15\textwidth} 
          \centering 
          \includegraphics[width=\linewidth]{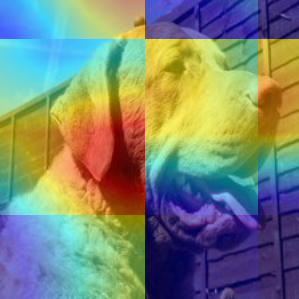}
          \caption*{\small Reshuffled Image}
    \end{minipage}
    \caption{The attention heatmaps of the raw images, shuffled images, and reshuffled heatmaps on the shuffled image generated on Inception-v3 model using Grad-CAM.}
    \label{fig:attention_bs}
\end{figure}

Adversarial attacks can generally be categorized into two types: white-box attacks and black-box attacks. White-box attacks involve having complete access to the target model's architecture and parameters. In contrast, black-box attacks only have limited information about the target model, which makes them more applicable for real-world applications. In the white-box setting, some studies employ the gradient with respect to the input sample to generate adversarial examples~\cite{goodfellow2015FGSM}. Those crafted adversarial examples exhibit transferability across neural models~\cite{Papernot2017blackbox}, which is the ability of adversarial examples generated on one model to deceive not only the victim model but also other models, making them suitable for black-box attacks. However, existing attack methods~\cite{kurakin2017adversarial,madry2018pgd} demonstrate outstanding white-box attack performance but relatively poorer transferability, limiting their efficacy in attacking real-world applications.

Recently, several approaches have emerged to enhance adversarial transferability, including incorporating momentum into gradient-based attacks~\cite{dong2018boosting,lin2020nesterov}, attacking multiple models simultaneously~\cite{liu2017delving}, transforming the image before gradient calculation~\cite{xie2019improving,wang2023structure}, leveraging victim model features~\cite{wang2021Feature}, and modifying the forward or backward process~\cite{wang2023diversifying,wang2023rethinking}. Among these, input transformation based methods, which modify the input image for gradient calculation, have demonstrated significant effectiveness in improving transferability.
However, we find that all the existing input transformation based attacks result in different attention heatmaps~\cite{selvaraju2017grad} on different models. This discrepancy in attention heatmaps could potentially limit the extent of adversarial transferability.

The attention heatmaps highlight the crucial regions for classification. Motivated by this, we aim to maintain consistency in the attention heatmaps of adversarial examples across different models. Since we only have access to a single white-box model for the attack, we initially explore methods to disrupt the attention heatmaps. As shown in Fig.~\ref{fig:attention_bs}, we can disrupt the intrinsic relation within the image by randomly shuffling the divided blocks of the image, leading to different attention heatmaps compared with the raw image. Based on this finding, we propose a novel input transformation based attack, called block shuffle and rotation (\name), which optimizes the adversarial perturbation on several transformed images to eliminate the variance among the attention heatmaps on various models. In particular, \name randomly divides the image into several blocks, which are subsequently shuffled and rotated to create new images for gradient calculation. To eliminate the variance of random transformation and stabilize the optimization, \name adopts the average gradient on several transformed images. 

In summary, we highlight our contributions as follows:

\begin{itemize}[leftmargin=*,noitemsep,topsep=2pt]
    \item We show that breaking the intrinsic relation of the image can disrupt the attention heatmaps of the deep model.
    \item We propose a new attack called block shuffle and rotation (\name), which is the first input transformation based attack to disrupt attention heatmaps for better transferability.
    \item Empirical evaluations on the ImageNet dataset demonstrate \name achieves much better transferability than the state-of-the-art input transformation based attacks.
    \item \name is compatible with other transfer-based attacks and can be integrated with each other to boost the adversarial transferability further.
\end{itemize}

\section{Related Work}
Here we briefly introduce adversarial attacks and defenses and summarize the visualization of attention heatmaps.
\subsection{Adversarial Attacks}
\citet{szegedy2014intriguing} first identified adversarial examples, which bring a great threat to DNN applications. Recently, numerous attacks have been proposed, which mainly fall into two categories: 1) \textit{white-box attacks} can access all the information of the target model, such as gradient, weight, architecture, \etc. Gradient-based attacks~\cite{madry2018pgd,carlini2017towards} that maximize the loss function using the gradient \wrt input are the predominant white-box attacks. 2) \textit{black-box attacks} are only allowed limited access to the target model, which can be further categorized into  \textit{Score-based attacks}~\cite{uesato2018adversarial,guo2019simple}, \textit{Decision-based attacks}~\cite{li2020qeba,wang2022triangle} and \textit{Transfer-based attacks}~\cite{dong2018boosting,xie2019improving,ge2023improving}. Among these, transfer-based attacks cannot access the target model, in which the attacker adopts the adversarial examples generated by the surrogate model to attack the target model directly. Hence, transfer-based attacks can be effectively deployed in the real world and have attracted wide interest.

Fast Gradient Sign Method (FGSM)~\cite{goodfellow2015FGSM} adds the perturbation in the gradient direction of the input. Iterative FGSM (I-FGSM)~\cite{kurakin2017adversarial} extends FGSM into an iterative version, showing better white-box attack performance but poor transferability. Recently, numerous works have been proposed to improve adversarial transferability.

MI-FGSM~\cite{dong2018boosting} introduces momentum into I-FGSM to stabilize the optimization procedure and escape the local optima. Later, more advanced momentum based attacks are proposed to further boost transferability, such as NI-FGSM~\cite{lin2020nesterov}, VMI-FGSM~\cite{wang2021enhancing}, EMI-FGSM~\cite{wang2021boosting}, PGN~\cite{ge2023boosting} and so on. Ensemble attacks~\cite{liu2017delving,xiong2022stochastic,li2020learning} generate more transferable adversarial examples by attacking multiple models simultaneously. Several works~\cite{zhou2018transferable,wu2020boosting} disrupt the feature space to generate adversarial examples.

On the other hand, input transformation has become one of the most effective ways to improve transferability. Diverse input method (DIM)~\cite{xie2019improving} first resizes the image into random size and adds the padding to fixed size before the gradient calculation. Translation invariant method (TIM)~\cite{dong2019evading} translates the image into a set of images for gradient calculation, which is approximated by convolving the gradient with a Gaussian kernel. Scale invariant method (SIM)~\cite{lin2020nesterov} calculates the gradient of several scaled images. \textit{Admix}~\cite{wang2021admix} mixes a small portion of images from other categories to the input image to craft a set of admixed images. PAM~\cite{zhang2023improving} augments the input images from several augmentation paths.



In this work, we propose a new input transformation, called \name, which breaks the intrinsic semantic relation of the input image for more diverse transformed images and results in much better transferability than the baselines.

\subsection{Adversarial Defenses}
With the increasing interest of adversarial attacks, researchers have been struggling to mitigate such threats. Adversarial training~\cite{goodfellow2015FGSM,madry2018pgd,tramer2018ensemble} injects adversarial examples into the training process to boost the model robustness. Among them, \citet{tramer2018ensemble} propose ensemble adversarial training using adversarial perturbation generated on several models, showing great effectiveness against transfer-based attacks. Denoising filter is a data preprocessing method that filters out the adversarial perturbation before feeding them into the target model. For instance, \citet{liao2018defense} design a high-level representation guided denoiser (HGD) based on U-Net to eliminate adversarial perturbation. \citet{naseer2020a} train a neural representation purifier (NRP) using a self-supervised adversarial training mechanism to purify the input sample. Researchers also introduce several input transformation based defenses that transform the image before prediction to eliminate the effect of adversarial perturbation, such as random resizing and padding~\cite{xie2019improving}, feature squeezing using bit reduction~\cite{xu2018bitred}, feature distillation~\cite{liu2019FD}. Different from the above empirical defenses, several certified defense methods provide a provable defense in a given radius, such as interval bound propagation (IBP)~\cite{gowal2019scalable}, CROWN-IBP~\cite{zhang2020towards}, randomized smoothing (RS)~\cite{Cohen2019RS}, \etc.

\subsection{Attention Heatmaps}
As the inner mechanism of deep models remains elusive to researchers, several techniques~\cite{smilkov2017smoothgrad,sundararajan2017axiomatic,selvaraju2017grad} have been developed to interpret these models. Among them, attention heatmap is a widely adopted way to interpret deep models. For instance, \citet{zhou2016cam} adopts global average pooling to highlight the discriminative object parts. \citet{selvaraju2017grad} proposed Grad-CAM using the gradient information to generate more accurate attention heatmaps. In this work, we adopt the attention heatmap to investigate how to boost adversarial transferability.

\begin{figure*}
    \centering
    \begin{minipage}[b]{0.12\textwidth} 
          \centering 
          \makebox[0pt][r]{\makebox[15pt]{\raisebox{25pt}{\rotatebox[origin=c]{90}{\small Source: Inc-v3}}}}%
          \includegraphics[width=\linewidth]{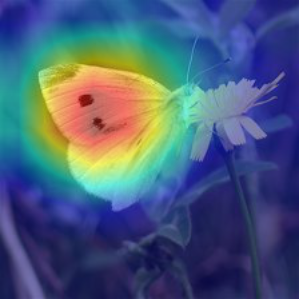}\\
          \vspace{0.1em}
          \makebox[0pt][r]{\makebox[15pt]{\raisebox{25pt}{\rotatebox[origin=c]{90}{\small Target: Inc-v4}}}}%
          \includegraphics[width=\linewidth]{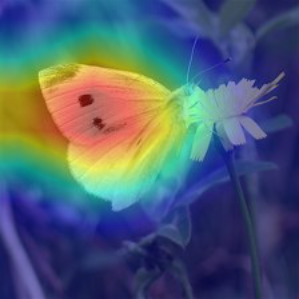}
          \vspace{-2em}
          \caption*{\small Raw Image}
    \end{minipage}
    \hspace{-0.2em}
    \begin{minipage}[b]{0.12\textwidth} 
          \centering 
          \includegraphics[width=\linewidth]{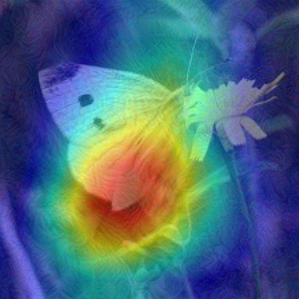}\\
          \vspace{0.1em}
          \includegraphics[width=\linewidth]{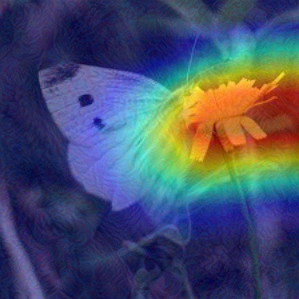}
          \vspace{-2em}
          \caption*{\small DIM}
    \end{minipage}
    \hspace{-0.2em}
    \begin{minipage}[b]{0.12\textwidth} 
          \centering 
          \includegraphics[width=\linewidth]{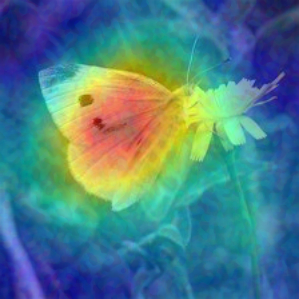}\\
          \vspace{0.1em}
          \includegraphics[width=\linewidth]{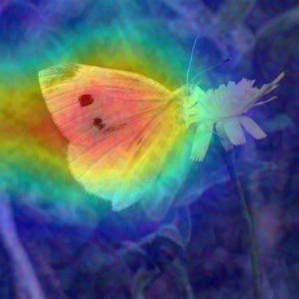}
          \vspace{-2em}
          \caption*{\small TIM}
    \end{minipage}
     \hspace{-0.2em}
    \begin{minipage}[b]{0.12\textwidth}     
          \centering 
          \includegraphics[width=\linewidth]{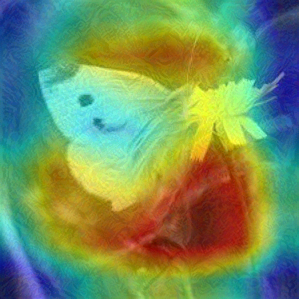}\\
          \vspace{0.1em}
          \includegraphics[width=\linewidth]{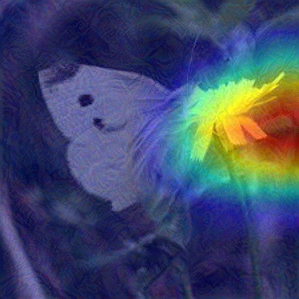}
          \vspace{-2em}
          \caption*{\small SIM}
    \end{minipage}
    \hspace{-0.2em}
    \begin{minipage}[b]{0.12\textwidth}    
          \centering 
          \includegraphics[width=\linewidth]{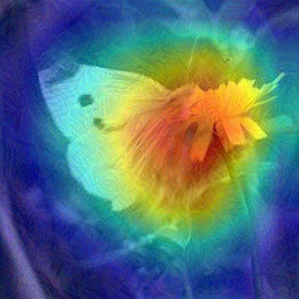}\\
          \vspace{0.1em}
          \includegraphics[width=\linewidth]{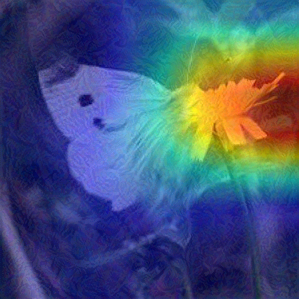}
          \vspace{-2em}
          \caption*{\small \textit{Admix}}
    \end{minipage}
    \hspace{-0.2em}
    \begin{minipage}[b]{0.12\textwidth}    
          \centering 
          \includegraphics[width=\linewidth]{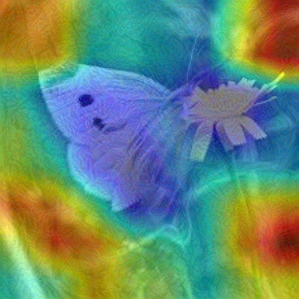}\\
          \vspace{0.1em}
          \includegraphics[width=\linewidth]{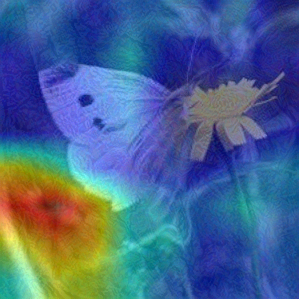}
          \vspace{-2em}
          \caption*{\small PAM}
    \end{minipage}
    \hspace{-0.2em}
    \hspace{-0.2em}
    \begin{minipage}[b]{0.12\textwidth} 
          \centering 
          \includegraphics[width=\linewidth]{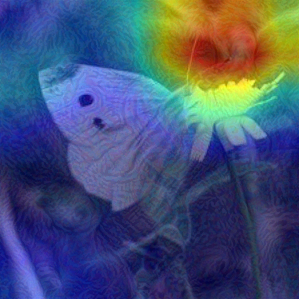}\\
          \vspace{0.1em}
          \includegraphics[width=\linewidth]{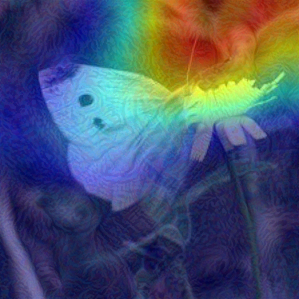}
          \vspace{-2em}
          \caption*{\small \name}
    \end{minipage}
   
    \caption{Attention heatmaps of adversarial examples generated by various input transformations using Grad-CAM.}
    \label{fig:attention_attack}

\end{figure*}
\section{Methodology}
In this section, we first introduce the preliminaries and motivation. Then we provide a detailed description of our \name, and summarize the difference between RLFAT~\cite{song2020rbs} and \name. 
\subsection{Preliminaries}
Given a victim model $f$ with parameters $\theta$ and a clean image $\mathbf{x}$ with ground-truth label $\mathbf{y}$, the attacker aims to generate an adversarial example $\mathbf{x}^{adv}$ that is indistinguishable from original image $\mathbf{x}$ (\ie, $\|\mathbf{x}^{adv} - \mathbf{x}\|_{p}\le \epsilon$) but can fool the victim model $f(\mathbf{x}^{adv};\theta)\neq f(\mathbf{x};\theta) = \mathbf{y}$. Here $\epsilon$ is the perturbation budget, and $\|\cdot\|_{p}$ is the $\ell_{p}$ norm distance. In this paper, we adopt $\ell_{\infty}$ distance to align with existing works. To generate such an adversarial example, the attacker often maximizes the objective function, which can be formalized as:
 \begin{equation}
    \mathbf{x}^{adv} = \argmax_{\|\mathbf{x}^{adv} - \mathbf{x}\|_{p}\le \epsilon} J(\mathbf{x}^{adv},\mathbf{y};\theta),
\end{equation}
where $J(\cdot)$ is the corresponding loss function (\eg, cross-entropy loss). 
For instance, FGSM~\cite{goodfellow2015FGSM} updates the benign sample by adding a small perturbation in the direction of the gradient sign:
\begin{equation}
    \mathbf{x}^{adv} = \mathbf{x} + \epsilon\cdot\operatorname{sign}(\nabla_{x}J(\mathbf{x},\mathbf{y};\theta)).
\end{equation}
FGSM can efficiently craft adversarial examples but showing poor attack performance. Thus, I-FGSM~\cite{kurakin2017adversarial} extends FGSM into an iterative version, which iteratively updates the the adversarial example by adding small perturbation with a step size $\alpha$:
\begin{gather}
    \mathbf{x}^{adv}_t = \mathbf{x}^{adv}_{t-1} + \alpha \cdot \operatorname{sign}(\nabla_{\mathbf{x}^{adv}_{t-1}}J(\mathbf{x}^{adv}_{t-1},\mathbf{y};\theta)),
\end{gather}
where $\mathbf{x}^{adv}_0 = \mathbf{x}$. 
Considering the poor transferability of I-FGSM, MI-FGSM~\cite{dong2018boosting} integrates momentum into the gradient for more transferable adversarial examples:
\begin{equation}
    \begin{aligned}
        g_t = \mu \cdot g_{t-1} + \frac{\nabla_{\mathbf{x}^{adv}_{t-1}}J(\mathbf{x}^{adv}_{t-1},\mathbf{y};\theta)}{\|\nabla_{\mathbf{x}^{adv}_{t-1}}J(\mathbf{x}^{adv}_{t-1},\mathbf{y};\theta)\|_1},\\
        \mathbf{x}^{adv}_t = \mathbf{x}^{adv}_{t-1} + \alpha \cdot \operatorname{sign}(g_{t}), \ g_0 = 0,
        \label{eq:mifgsm}
    \end{aligned}    
\end{equation}
where $\mu$ is the decay factor. Suppose $\mathcal{T}$ is a transformation operator, existing input transformation based attacks are often integrated into MI-FGSM to boost adversarial transferability, \ie, adopting $\nabla_{\mathbf{x}^{adv}_{t-1}}J(\mathcal{T}(\mathbf{x}^{adv}_{t-1}),\mathbf{y};\theta)$ for Eq.~\eqref{eq:mifgsm}.

\subsection{Motivation}

 

While different models may have distinct parameters and architectures, there are often shared characteristics in their learned features for image recognition tasks~\cite{zhou2016cam,smilkov2017smoothgrad}. In this work, we hypothesize that the adversarial perturbations which target these salient features have a greater impact on adversarial transferability. \citet{wu2020boosting} find disrupting the attention heatmaps can enhance transferability, which also supports our hypothesis. Intuitively, when the attention heatmaps of adversarial examples exhibit consistency across various models, it is expected to yield better adversarial transferability. To explore this idea, we initially assess the consistency of attention heatmaps generated by Grad-CAM~\cite{Selvaraju2017gradcam} for several input transformation based attacks. Unfortunately, as shown in Fig.~\ref{fig:attention_attack}, the attention heatmaps of adversarial examples on the white-box model are different from that on the target black-box model, resulting in limited adversarial transferability. This finding motivates us to investigate a new problem:

\textit{How can we generate adversarial examples with consistent attention heatmaps across different models?}

To maintain the consistency of attention heatmaps across multiple models, one direct approach is to optimize the adversarial perturbation by utilizing gradients \wrt the input image obtained from different models, \aka ensemble attack~\cite{liu2017delving}. By incorporating gradients from various models, each associated with its own attention heatmap for the same input image, ensemble attack helps eliminate the variance among attention heatmaps, resulting in improved transferability. In practice, however, it is often challenging and costly to access multiple models, making it more feasible to work with a single surrogate model. In this work, we explore how to obtain the gradients with different attention heatmaps on a single model using input transformation. With such transformation, we can optimize the perturbation to eliminate the variance among attention heatmaps of various transformed images, thus enhancing the consistency of attention heatmaps and adversarial transferability.


\subsection{Block Shuffle and Rotation}
With a single source model, we have to transform the input image to obtain diverse attention heatmaps when calculating the gradient. Thus, we should address the problem:

\textit{How to transform the image to disrupt the attention heatmap on a single source model?}

Attention heatmap highlights the significant features that contribute to the deep model's accurate prediction. For human perception, we are capable of recognizing objects based on partial visual cues, even when they are partially obstructed by other objects. For instance, we can identify a cat by observing only a portion of its body (\eg, head). This observation motivates us to employ image transformations that draw attention to specific regions of the main object, thereby varying the attention heatmap on a single model. Intuitively, randomly masking the object partially can force the deep model to focus on the remaining object, leading to various attention heatmaps. However, masking the object leads to a loss of information in the image, rendering the gradient meaningless for the masked block. Consequently, it can slow down the attack efficiency and effectiveness.
\begin{algorithm}[tb]
    \algnewcommand\algorithmicinput{\textbf{Input:}}
    \algnewcommand\Input{\item[\algorithmicinput]}
    \algnewcommand\algorithmicoutput{\textbf{Output:}}
    \algnewcommand\Output{\item[\algorithmicoutput]}

    \caption{Block Shuffle and Rotation}
    \label{alg:BSR}
	\begin{algorithmic}[1]
		\Input A classifier $f$ with parameters $\theta$, loss function $J$; a raw example $x$ with ground-truth label $y$; the magnitude of perturbation $\epsilon$; number of iteration $T$; decay factor $\mu$; the number of transformed images $N$; the number of blocks $n$; the maximum angle $\tau$ for rotation
        \Output An adversarial example $x^{adv}$
		\State $\alpha = \epsilon/T$, $g_0 = 0$
		\For{$t = 1 \rightarrow T$}
		    \State Generate several transformed images: $\mathcal{T}(\mathbf{x}^{adv},n,\tau)$
		    \State Calculate the average gradient $\bar{g}_{t}$ by Eq.~\eqref{eq:gradient}:
		    \State Update the momentum $g_{t}$ by:
		    \begin{equation*}
		        g_{t} = \mu \cdot g_{t-1} + \frac{\bar{g}_{t}}{\|\bar{g}_{t}\|_1}
		    \end{equation*}
		    \State Update the adversarial example:
		    \begin{equation}
		        x^{adv}_{t} = x^{adv}_{t-1} + \alpha \cdot \operatorname{sign}(g_{t})
		    \end{equation}
		\EndFor
        \State \Return $x^{adv}_{T}$
	\end{algorithmic} 
\end{algorithm} 

On the other hand, humans exhibit a remarkable ability to not only recognize the visible parts of the objects but also mentally reconstruct the occluded portions when the objects are partially obstructed by other elements. This cognitive process is attributed to our perception of intrinsic relationships inherent within the object, such as the understanding that a horse's legs are positioned beneath its body. Recognizing the significance of intrinsic relationships for human perception, we try to disrupt these relationships to affect attention heatmaps. In particular, we split the image into several blocks and shuffle the blocks to construct new images that appear visually distinct from the original one. As expected, the attention heatmaps are also disrupted on the transformed image, even when recovering the attention heatmaps to match the original image, as depicted in Fig.~\ref{fig:attention_bs}. Thus, we can break the intrinsic relationships for more diverse attention heatmaps to boost adversarial transferability.

To achieve this goal, we propose a new input transformation $\mathcal{T}(\mathbf{x}, n, \tau)$, which randomly splits the image into $n\times n$ blocks followed by the random shuffling of these blocks. To further disrupt the intrinsic relationship, each block is independently rotated by an angle within the range of $-\tau \le \beta \le \tau$ degrees. During the rotation of each block, any portions that extend beyond the image boundaries are removed while the resulting gaps are filled with zero. 

With the transformation $\mathcal{T}(\mathbf{x}, n, \tau)$, we can obtain more diverse attention heatmaps compared to those obtained from various models. Hence, strengthening the consistency of heatmaps (\textit{harder}) helps result in consistent attention across models (\textit{easier}). This motivates us to adopt transformation $\mathcal{T}(\mathbf{x}, n, \tau)$ for gradient calculation to achieve more consistent heatmaps. Note that transformation $\mathcal{T}(\mathbf{x}, n, \tau)$ cannot guarantee that all transformed images are correctly classified ($\sim 86.8\%$ on Inc-v3). To eliminate the variance introduced by the intrinsic relation corruption, we calculate the average gradient on $N$ transformed images as follows:
\begin{equation}
    \bar{g} = \frac{1}{N} \sum_{i=0}^{N} \nabla_{\mathbf{x}^{adv}} J(\mathcal{T}(\mathbf{x}^{adv},n,\tau),\mathbf{y};\theta)).
    \label{eq:gradient}  
\end{equation}

The input transformation $\mathcal{T}(\mathbf{x}, n, \tau)$ is general to any existing attacks. Here we integrate it into MI-FGSM, denoted as Block Shuffle and Rotation (\name), and summarize the algorithm in Algorithm~\ref{alg:BSR}.

\subsection{\name Vs. RLFAT}
\citet{song2020rbs} propose Robust Local Features for Adversarial Training (RLFAT) by minimizing the distance between the high-level feature of the original image and block shuffled image for better generalization. Since RLFAT also shuffles the image blocks, we highlight the difference between \name and RLFAT as follows:
\begin{itemize}
    \item \textbf{Goal.} \name aims to generate more transferable adversarial examples, while RLFAT boosts the generalization of adversarial training.
    \item \textbf{Strategy.} \name shuffles and rotates the image blocks, while RLFAT only shuffles the blocks.
    \item \textbf{Usage.} \name directly adopts the transformed images for gradient calculation to craft adversarial examples, while RLFAT treats it as a regularizer for the original image.
\end{itemize}
With the different goals, strategies, and usages, BSR should definitely be a new and novel input transformation based attack to effectively boost the adversarial transferability.
\section{Experiments}
In this section, we conduct empirical evaluations on ImageNet dataset to evaluate the effectiveness of \name.

\begin{table*}
\begin{center}

\begin{tabular}{c>{\rowmac}c>{\rowmac}c>{\rowmac}c>{\rowmac}c>{\rowmac}c>{\rowmac}c>{\rowmac}c>{\rowmac}c}
\toprule
Model & Attack & Inc-v3 & Inc-v4 & IncRes-v2 & Res-101 & Inc-v3$_{ens3}$ & Inc-v3$_{ens4}$ & IncRes-v2$_{ens}$\\
\midrule
\multirow{6}{*}{Inc-v3}& DIM & ~~98.6* & 64.4 & 60.2 & 53.5 & 18.8 & 18.4 &  9.5\\
& TIM & \textbf{100.0}* & 49.3 & 43.9 & 40.2 & 24.6 & 21.7 & 13.4\\
& SIM & \textbf{100.0}* & 69.5 & 68.5 & 63.5 & 32.3 & 31.0 & 17.4\\
& \textit{Admix} & \textbf{100.0}* & 82.2 & 81.1 & 73.8 & 38.7 & 37.9 & 19.8 \\
& PAM  & \textbf{100.0}*           &76.4  &75.5  &69.6  &39.0  &38.8  &20.0\\
& \name & \setrow{\bfseries}100.0* & 96.2 & 94.7 & 90.5 & 55.0 & 51.6 & 29.3 \clearrow\\
\midrule
\multirow{6}{*}{Inc-v4}& DIM & 72.0 & ~~97.6* & 63.8 & 57.2 & 22.6 & 21.1 & 11.7\\
& TIM & 59.1 & ~~99.7* & 49.0 & 41.9 & 26.8 & 22.9 & 16.6\\
& SIM & 80.4 & ~~99.7* & 73.4 & 69.4 & 48.6 & 45.2 & 29.6\\
& \textit{Admix} & 89.0 & ~~\textbf{99.9}* & 85.3 & 79.0 & 55.5 & 51.7 & 32.3\\
& PAM  & 86.7          &~~\textbf{99.9}*  &81.6  &75.9  &55.4 &50.5  &33.2\\
& \name & \setrow{\bfseries}96.1  & ~~99.9* & 93.4 & 88.4 & 57.6 & 52.1 \clearrow & \textbf{34.3} \\
\midrule
\multirow{6}{*}{IncRes-v2}& DIM & 70.3 & 64.7 & ~~93.1* & 58.0 & 30.4 & 23.5 & 16.9\\
& TIM & 62.2 & 55.6 & ~~97.4* & 50.3 & 32.4 & 27.5 & 22.6\\
& SIM & 85.9 & 80.0 & ~~98.7* & 76.1 & 56.2 & 49.1 & 42.5\\
& \textit{Admix} & 90.8 & 86.3 & ~~\textbf{99.2}* & 82.2 & 63.6 & 56.6 & 49.4\\
& PAM & 88.6&86.3&~~99.4*&81.6&66.0&58.3&\textbf{51.0}\\
& \name &  \setrow{\bfseries}94.6& 93.8\clearrow & ~~98.5* & \setrow{\bfseries}90.7 & 71.4 & 63.1 & 51.0 \clearrow\\
\midrule
\multirow{6}{*}{Res-101}& DIM & 76.0 & 68.4 & 70.3 & ~~98.0* & 34.7 & 31.8 & 19.6\\
& TIM & 59.9 & 52.2 & 51.9 & ~~99.2* & 34.4 & 31.2 & 23.7\\
& SIM & 74.1 & 69.6 & 69.1 & ~~99.7* & 42.8 & 39.6 & 25.7\\
& \textit{Admix} & 84.5 & 80.2 & 80.7 & ~~\textbf{99.9}* & 51.6 & 44.7 & 29.9\\
& PAM & 77.4 & 73.9 & 75.7 & ~\textbf{99.9}* & 51.2 & 46.3 & 32.2\\
& \name &  \textbf{97.1}& \textbf{96.6} & \textbf{96.6} & ~~99.7* & \textbf{78.7} & \textbf{74.7} & \textbf{55.6} \\

\bottomrule
\end{tabular}
\caption{Attack success rates (\%) on seven models under single model setting with various single input transformations. The adversaries are crafted on Inc-v3, Inc-v4, IncRes-v2 and Res-101 respectively. * indicates white-box attacks.}
\label{tab:single_transformation}
\end{center}

\end{table*}
\begin{table*}
\begin{center}
\begin{tabular}{cccccccc}
\toprule
Attack  & Inc-v4 & IncRes-v2 & Res-101 & Inc-v3$_{ens3}$ & Inc-v3$_{ens4}$ & IncRes-v2$_{ens}$\\
\midrule
BSR-DIM  & 98.3\imp{31.9} & 94.8\imp{34.6} & 90.1\imp{36.6} & 57.8\imp{39.0} & 54.5\imp{36.1} & 32.3\imp{22.8} \\
BSR-TIM & 94.7\imp{45.4} & 92.5\imp{48.6} & 87.0\imp{47.0} & 71.3\imp{46.7} & 68.4\imp{46.7} & 47.9\imp{34.5} \\
BSR-SIM & 99.4\imp{29.9}&98.4\imp{29.9}&97.8\imp{34.3}&84.3\imp{52.0}&81.4\imp{50.4}&59.2\imp{41.8}\\
BSR-\textit{Admix} & 98.9\imp{16.7}&98.8\imp{17.7}&98.2\imp{24.4}&89.1\imp{50.4}&86.9\imp{49.5}&68.0\imp{48.2}\\
BSR-PAM & 98.5\imp{22.1}&97.3\imp{21.8}&96.9\imp{27.3}&79.4\imp{40.4}&75.3\imp{36.5}&50.5\imp{30.5}\\
\midrule
\textit{Admix}-TI-DIM & 90.4 & 87.3 & 83.7 & 72.4 & 68.4 & 53.4\\
PAM-TI-DIM & 89.3 & 85.5 & 80.7 & 73.6 & 69.1 & 52.1\\
 BSR-TI-DIM&95.2&92.9&87.9&74.2&70.7&50.0\\
BSR-SI-TI-DIM&\textbf{98.5}&\textbf{97.1}&\textbf{95.4}&\textbf{90.6}&\textbf{90.0}&\textbf{75.1}\\
\bottomrule
\end{tabular}
\caption{Attack success rates (\%) on seven models under single model setting with various input transformations combined with \name. The adversaries are crafted on Inc-v3. \textcolor{Red}{$\uparrow$} indicates the increase of attack success rate when combined with \name.}
\label{tab:combined}
\end{center}
\vspace{-1em}
\end{table*}
\subsection{Experimental Setup}
\textbf{Dataset.} We evaluate our proposed \name on 1000 images belonging to 1000 categories from the validation set of ImageNet dataset~\cite{Russa2015imagenet}.

\textbf{Models.} We adopt four popular models, \ie, Inception-v3 (Inc-v3)~\cite{szegedy2016inceptionv3}, Inception-v4 (Inc-v4), Inception-Resnet-v3 (IncRes-v3)~\cite{szegedy2017inceptionv4}, Resnet-v2-101 (Res-101)~\cite{he2016resnet}, and three ensemble adversarially trained models, \ie, $\text{Inc-v3}_{ens3}$, $\text{Inc-v3}_{ens4}$, $\text{IncRes-v2}_{ens2}$~\cite{tramer2018ensemble} as victim models to evaluate the transferability. To further verify the effectiveness of \name, we utilize several advanced defense methods, including HGD~\cite{liao2018defense}, R\&P~\cite{xie2018mitigating}, NIPS-r3\footnote{https://github.com/anlthms/nips-2017/tree/master/mmd}, Bit-RD~\cite{xu2018bitred}, JPEG~\cite{guo2018JPEG}, FD~\cite{liu2019FD}, RS~\cite{Cohen2019RS} and NRP~\cite{naseer2020a}.

\textbf{Baselines.} To verify the effectiveness of \name, we choose five competitive input transformation based attacks as our baselines, \ie DIM~\cite{xie2019improving}, TIM~\cite{dong2019evading}, SIM~\cite{lin2020nesterov}, \textit{Admix}~\cite{wang2021admix} and PAM\cite{zhang2023improving}. For fairness, all the input transformations are integrated into MI-FGSM~\cite{dong2018boosting}.

\textbf{Parameters Settings.} We set the maximum perturbation $\epsilon = 16$, number of iteration $T = 10$, step size $\alpha = \epsilon/T$ and the decay factor $\mu=1$ for MI-FGSM~\cite{dong2018boosting}. DIM~\cite{xie2019improving} adopts the transformation probability of $0.5$. TIM~\cite{dong2019evading} utilizes a kernel size of $7\times 7$. The number of copies of SIM~\cite{lin2020nesterov} and \textit{Admix}~\cite{wang2021admix} is $5$. \textit{Admix} admixes $3$ images with the strength of $0.2$. The number of scale for PAM~\cite{zhang2023improving} is 4, and the number of augmented path is 3 . Our \name splits the image into $2\times2$ blocks with the maximum rotation angle $\tau=24^\circ$ and calculates the gradients on $N=20$ transformed images.

\subsection{Evaluation on Single Model}
We first evaluate the attack performance on various input transformation based attacks, \ie, DIM TIM, SIM, \textit{Admix}, PAM, and our proposed \name. We craft the adversaries on the four standard trained models and test them on seven models. The attack success rates, \ie, the misclassification rates of the victim model on the adversarial examples, are summarized in Tab.~\ref{tab:single_transformation}. Each column denotes the model to be attacked and each row indicates that the attacker generates the adversarial examples on the corresponding models.

\begin{table*}
\begin{center}
\begin{tabular}{cccccccc}
\toprule
Attack & Inc-v3 & Inc-v4 & IncRes-v2 & Res-101 & Inc-v3$_{ens3}$ & Inc-v3$_{ens4}$ & IncRes-v2$_{ens}$\\
\midrule
DIM & ~~99.0* & 97.1* & ~~93.4* & ~~99.7* & 57.6 & 51.5 & 35.9\\
TIM & ~~99.8* & 97.4* & ~~94.7* & ~~99.8* & 61.6 & 55.5 & 45.6\\
SIM & ~~99.9* & 99.1* & ~~98.5* & \textbf{100.0}* & 78.4 & 75.2 & 60.6\\
\textit{Admix} & \textbf{100.0}* & 99.6* & ~~99.0* & \textbf{100.0}* & 85.1 & 80.9 & 67.8\\
PAM &~~99.9*&99.7*&~~99.4*&\textbf{100.0}*&86.1&81.6& 69.1\\
\name & \textbf{100.0}* & \textbf{99.9}* & ~~99.9* & ~~99.9* & \textbf{92.4} & \textbf{89.0} & \textbf{77.2} \\
\midrule
 \textit{Admix}-TI-DIM & ~~99.6* & 98.8* & ~~98.2* & ~~99.8* & 93.1 & 92.4 & 89.4 \\
 PAM-TI-DIM&~~99.8*&99.8*&~~99.2*&~~\textbf{99.8}*&95.8&95.2&93.0\\
 \cname{-TI-DIM} & ~~99.8* & 99.8* & ~~99.7* & ~~\textbf{99.8}* & 96.1 & 95.1 & 90.8 \\
 \cname{SI-TI-DIM} &~~\textbf{99.9}*&\textbf{99.9}*&~~\textbf{99.9}*&~~\textbf{99.8}*&\textbf{99.1}&\textbf{99.1}&\textbf{97.0}\\
\bottomrule
\end{tabular}

\caption{Attack success rates (\%) on seven models under ensemble model setting with various input transformations. The adversaries are crafted on Inc-v3, Inc-v4, IncRes-v2 and Res-101 model. * indicates white-box attacks.}
\label{tab:ensemble}
\end{center}

\end{table*}
\begin{table*}
\begin{center}

\begin{tabular}{lccccccccc}
\toprule
Method & HGD & R\&P & NIPS-r3 & Bit-RD &JPEG & FD & RS & NRP & Average\\
\midrule
\textit{Admix}-TI-DIM & 92.8&93.5&94.5&82.4&97.6&90.9&72.6 &80.4&88.1\\
PAM-TI-DIM&95.4&95.3&96.4&85.9&98.4&93.4&74.0&83.8&91.6\\
BSR-TI-DIM&97.1&98.0&97.9&84.9&98.8&93.2&69.1&73.6&89.1\\
BSR-SI-TI-DIM& \textbf{98.5}& \textbf{99.1}&\textbf{99.4}&\textbf{91.4}&\textbf{99.2}&\textbf{97.1}&\textbf{83.9}&\textbf{84.2}&\textbf{94.1}\\
\bottomrule
\end{tabular}

\caption{Attack success rates (\%) of eight defense methods by \textit{Admix}, SSA and \name input transformations. The adversaries are crafted on Inc-v3, Inc-v4, IncRes-v2 and Res-101 synchronously.}
\vspace{-1em}
\label{tab:defense}
\end{center}
\end{table*}
It can be observed that for DIM and TIM, DIM exhibits superior performance on standard trained models while TIM exhibits better transferability on adversarially trained models. SIM, as a special case of \textit{Admix}, can achieve better performance than DIM and TIM, while \textit{Admix} shows the best performance on standardly trained model among the four baselines and PAM exhibits better transferability on adversarially trained model. In contrast, our proposed \name, surpasses existing input transformation based attacks while maintaining comparable performance in white-box attacks. Remarkably, on standard trained models, \name attains an average attack success rate of $93.8\%$, exhibiting a substantial improvement over \textit{Admix} by a clear margin of at least $6.5\%$. Similarly, on adversarially trained models, \name achieves an average attack success rate of $56.2\%$, outperforming PAM by a significant margin of $11.0\%$. These exceptional findings substantiate the superiority of \name in generating transferable adversarial examples, thereby highlighting the importance of maintaining attention heatmap consistency across different models as a means to enhance transferability.



\subsection{Evaluation on Combined Input Transformation}
\label{sec:com}
Previous works~\cite{wang2021admix} have shown that a good input transformation based attack should not only exhibit better transferability, but also be compatible with other input transformations to generate more transferable adversarial examples. Following the evaluation setting of \textit{Admix}, we combine our \name with various input transformations, denoted \cname{DIM}, \cname{TIM}, \cname{SIM}, \cname{\textit{Admix}} and \cname{PAM}. Here we also combine \name with multiple input transformations, denoted as \cname{TI-DIM} and \cname{SI-TI-DIM} to compare \textit{Admix}-TI-DIM and PAM-TI-DIM. 

We report the attack success rates of the adversarial examples generated on Inc-v3 in Tab.~\ref{tab:combined} and the results for other models in Appendix. Our \name significantly improves the transferability of these input transformation based attacks. In general, \name can improve the attack success rate with a range from $16.7\%$ to $52.0\%$. In particular, when combining these input transformation with \name, the attack performance on adversarially trained models are significantly improved by a margin of $22.8\% \thicksim 52.0\%$. Although \textit{Admix}-TI-DIM demonstrates the best performance among combined methods, our proposed \cname{TI-DIM} surpasses \textit{Admix}-TI-DIM with a clear margin of $2.6\%$ on average attack success rates. Notably, when \name is combined with SI-DI-TIM, it further enhances transferability by a margin ranging from $8.1\%$ to $31.6\%$. This further supports the high effectiveness of \name and shows its excellent compatibility with other input transformation based attacks.

\begin{figure*}[t]
    \begin{minipage}[b]{.3\textwidth} 
        \centering
        \includegraphics[width=\linewidth]{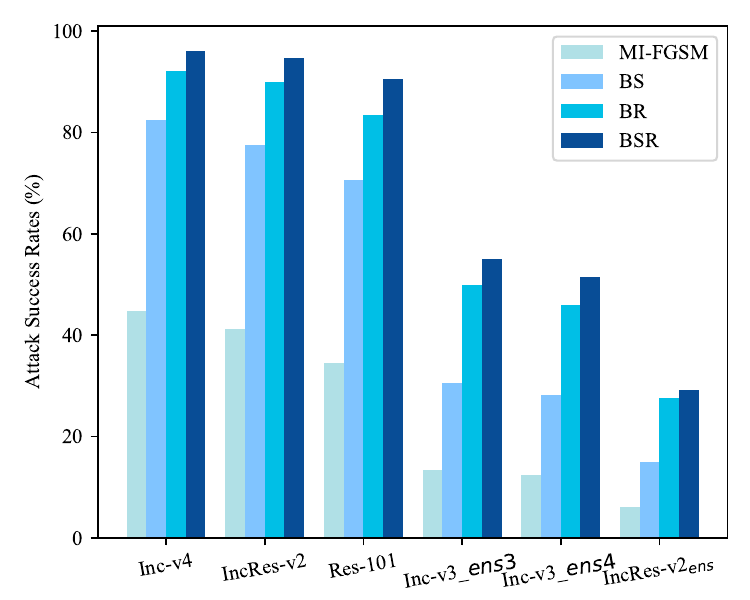}
        \caption{Attack success rates (\%) of various models on the adversarial examples generated by MI-FGSM, BS, BR and \name, respectively. }
        \label{fig:ablation}
    \end{minipage}
    \hspace{1em}
    \begin{minipage}[b]{.7\textwidth} 
        \centering
            \includegraphics[width=\linewidth]{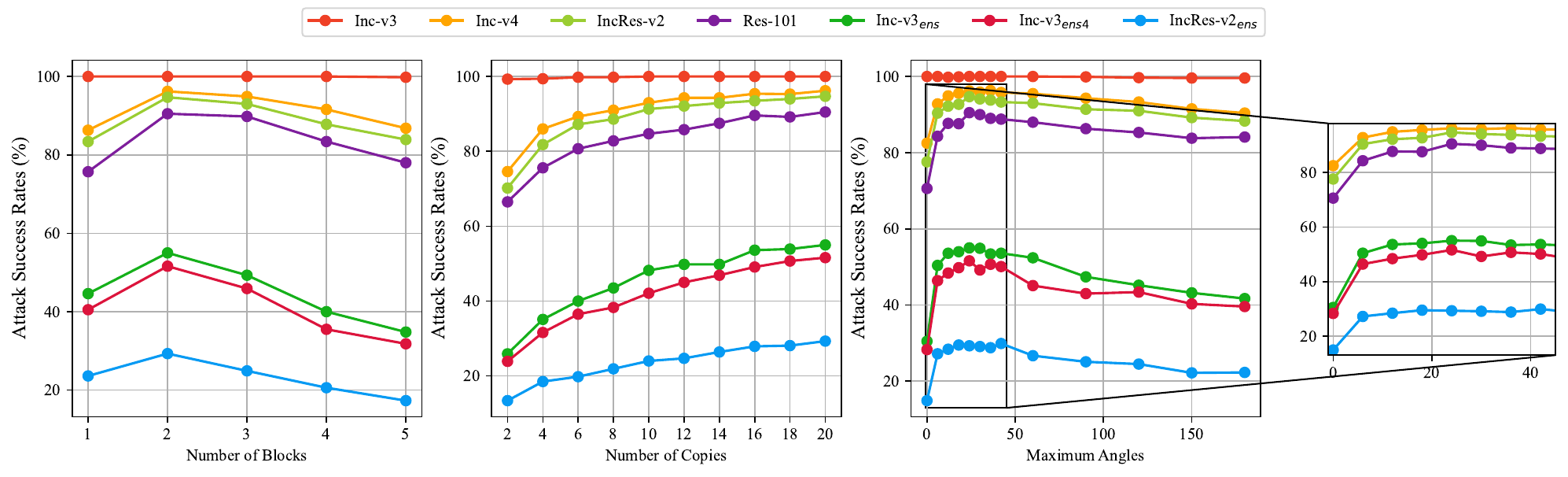}
            \label{fig:abl:n_block}

    \caption{Attack successful rates (\%) of various models on the adversarial examples generated by \name with various numbers of blocks, number of transformed images and range of the rotation angles. The adversarial examples are crafted on Inc-v3 model and tested on the other six models under the black-box setting.}
    \label{fig:para}
    \end{minipage}
\end{figure*}

\subsection{Evaluation on Ensemble Model}
\label{sec:ens}
Liu~\etal~\cite{liu2017delving} first propose ensemble attack to boost the transferability by synchronously attacking several models. To evaluate the compatibility of the proposed \name with ensemble attack, we generate the adversarial examples on four standard trained models and test them on adversarially trained models following the setting in MI-FGSM~\cite{dong2018boosting}. We evaluate the attack performance of single input transformation as well as \name combined with the existing input transformations, respectively.

As shown in Tab.~\ref{tab:ensemble}, under ensemble model setting, PAM showcases superior white-box attack capabilities, surpassing other baselines significantly with a margin of at least $0.7\%$ for adversarially trained models. In contrast, \name consistently outperforms PAM on these models by a margin of $6.3\%$ to $8.1\%$, and maintains comparable white-box attack performance with \textit{Admix}. Although PAM-TI-DIM can achieve at least $93.0\%$ attack success rate on adversarially trained models. Notably, \cname{SI-TI-DIM} achieves average attack success rate of $98.4\%$ on adversarially trained models, surpassing PAM-TI-DIM with a margin of at least $3.1\%$. Such remarkable attack performance validates the remarkable effectiveness of \name for improving the transferability and poses a huge threat to security-critical applications once again.

\subsection{Evaluation on Defense Method}
To thoroughly evaluate the effectiveness of our proposed method, we assess the attack performance of \name against several defense mechanisms, including HGD, R\&P, NIPS-r3, Bit-RD, JPEG, FD, RS and NRP. From previous experiments, combined input transformations with ensemble attack exhibit the best attack performance. Here we adopt the adversarial examples generated by PAM-TI-DIM, \cname{TI-DIM} and \cname{SI-TI-DIM} under the ensemble setting to attack these defense approaches.

As shown in Tab.~\ref{tab:defense}, \cname{SI-TI-DIM} achieves the average attack success rate of $94.1\%$, which outperforms PAM-TI-DIM by an average margin of $2.5\%$. Notably, even on the certified defense RS and powerful denoiser NRP, \cname{SI-TI-DIM} achieve the attack success rate of $83.9\%$ and $84.2\%$, which outperforms PAM-TI-DIM with a clear margin of $9.9\%$ and $0.4\%$, respectively. Such excellent attack performance further shows the superiority of \name and reveals the inefficiency of existing defenses.


\subsection{Ablation Study}
\label{sec:exp:ablation}
To further gain insight into the performance improvement of \name, we conduct ablation and hyper-parameter studies by generating the adversarial examples on Inc-v3 and evaluating them on the other six models. 

\textbf{On the effectiveness of shuffle and rotation.} After splitting the image into several blocks, we shuffle the image blocks and rotate each block. To explore the effectiveness of shuffle and rotation, we conduct two additional attacks, \ie, block shuffle (BS) and block rotation (BR). As shown in Fig.~\ref{fig:ablation}, BS and BR can achieve better transferability than MI-FGSM, supporting our proposition that optimizing the adversarial perturbation on the input image with different attention heatmaps can eliminate the variance among attention heatmaps to boost the transferability.  By combining BS and BR, \name exhibits the best transferability, showing its rationality and high effectiveness in crafting transferable adversarial examples.

\textbf{On the number of blocks $n$.} As shown in Fig.~\ref{fig:para}, when $n=1$, \name only rotates the raw image, which cannot bring disruption on the attention heatmaps, showing the poorest transferability. When $n>3$, increasing $n$ results in too much variance that cannot be effectively eliminated, decreasing the transferability. Hence, a suitable magnitude of disruption on the raw image is significant to improve the transferability and we set $n=2$ in our experiments.

\textbf{On the number of transformed images $N$.} Since \name introduces variance when breaking the intrinsic relation, we adopt the average gradient on $N$ transformed images to eliminate such variance. As shown in Fig.~\ref{fig:para}, when $N=5$, \name can already achieve better transferability than MI-FGSM, showing its high efficiency and effectiveness. When we increase $N$, the attack performance would be further improved and be stable when $N>20$. Hence, we set $N=20$ in our experiments.

\textbf{On the range of rotation angles $\tau$.} We randomly rotate the image blocks with the angle $-\tau \le \beta \le \tau$, which also affects the magnitude of disruption on the image. As shown in Fig.~\ref{fig:para}, we conduct the experiments from $\tau=6^\circ$ to $\tau=180^\circ$. When $\tau$ is smaller than $24^\circ$, increasing $\tau$ results in more disruption on the image so as to achieve better transferability. If we continue to increase $\tau$, the rotation will introduce too much disruption, which decays the performance. Hence, we set $\tau=24^\circ$ in our experiments.



\section{Conclusion}
Intuitively, the consistent attention heatmaps of adversaries on different models will have better transferability. However, we find that the existing input transformation based attacks often result in inconsistent attention heatmaps on various models, limiting the transferability. To this end, we propose a novel input transformation based attack called block shuffle and rotation (\name), which optimizes the perturbation on several transformed images with different attention heatmaps to eliminate the variance among the attention heatmaps on various models. Empirical evaluations on ImageNet dataset show that \name achieves better transferability than the SOTA attacks under various attack settings. We hope our approach can provide new insight to improve the transferability by generating adversarial examples with more stable attention heatmaps on different models.

{
    \small
    \bibliographystyle{ieeenat_fullname}
    \bibliography{main}
}
\newpage
~~
\newpage

\section{Appendix}
\label{sec:appendix}
In this appendix, we first provide more results of integrating \name with DIM, TIM, Admix and PAM, the adversaries are crafted on three other models \ie Inc-v4, IncRes-v2 and Res-101. Then we provide more comparison on the heatmap of various images using different methods.

\subsection{More Evaluations on Combined Input Transformation}
\label{sec:appendix_combined}
We provide the attack results of adversarial examples generated by integrating \name with DIM, TIM, \textit{Admix}, and PAM on the other three models, namely Inc-v4, IncRes-v2 and Res-101. As shown in Tab.~\ref{tab:combined_incv4}-\ref{tab:combined_res}, \name can remarkably boost the transferability of various input transformation based attacks when generating the adversarial examples on the three models.
\citet{zhang2023improving} have demonstrated that the combination of PAM-TI-DIM achieves state-of-the-art transferability. However, its performance on adversarially trained models remains relatively weak. Our proposed \cname{SI-DI-TIM} consistently outperforms PAM-TI-DIM with a margin of at least $6.4\%$. Such excellent performance improvement further validates the high teffectiveness of the proposed \name for boosting the adversarial transferability.





\subsection{Visualization on Attention Heatmaps}
\begin{figure*}
    \centering

    \begin{minipage}[b]{0.12\textwidth} 
          \centering
          \makebox[0pt][r]{\makebox[15pt]{\raisebox{25pt}{\rotatebox[origin=c]{90}{\small Source: Inc-v3}}}}%
          \includegraphics[width=\linewidth]{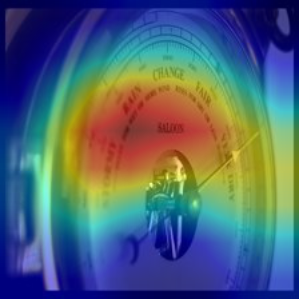}\\
          \vspace{0.1em}
          \makebox[0pt][r]{\makebox[15pt]{\raisebox{25pt}{\rotatebox[origin=c]{90}{\small Target: Inc-v4}}}}%
          \includegraphics[width=\linewidth]{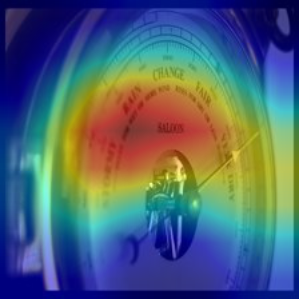}\\
         \vspace{0.1em}
          \makebox[0pt][r]{\makebox[15pt]{\raisebox{25pt}{\rotatebox[origin=c]{90}{\small Source: Inc-v3}}}}%
          \includegraphics[width=\linewidth]{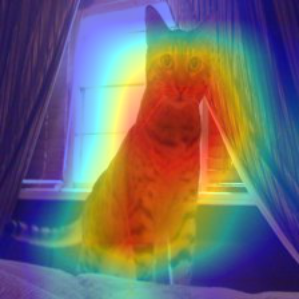}\\
          \vspace{0.1em}
          \makebox[0pt][r]{\makebox[15pt]{\raisebox{25pt}{\rotatebox[origin=c]{90}{\small Target: Inc-v4}}}}%
          \includegraphics[width=\linewidth]{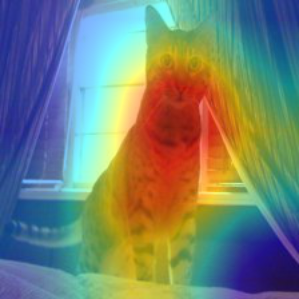}\\
          \vspace{0.1em}
          \makebox[0pt][r]{\makebox[15pt]{\raisebox{25pt}{\rotatebox[origin=c]{90}{\small Source: Inc-v3}}}}%
          \includegraphics[width=\linewidth]{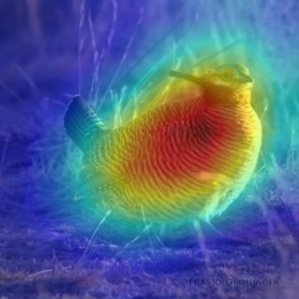}\\
          \vspace{0.1em}
          \makebox[0pt][r]{\makebox[15pt]{\raisebox{25pt}{\rotatebox[origin=c]{90}{\small Target: Inc-v4}}}}%
          \includegraphics[width=\linewidth]{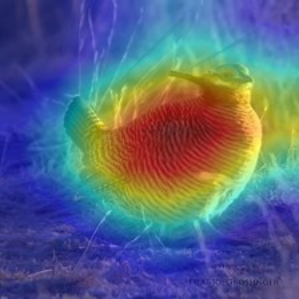}\\
          \vspace{0.1em}
          \makebox[0pt][r]{\makebox[15pt]{\raisebox{25pt}{\rotatebox[origin=c]{90}{\small Source: Inc-v3}}}}%
          \includegraphics[width=\linewidth]{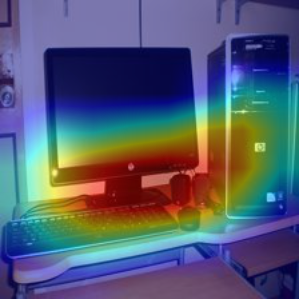}\\
          \vspace{0.1em}
          \makebox[0pt][r]{\makebox[15pt]{\raisebox{25pt}{\rotatebox[origin=c]{90}{\small Target: Inc-v4}}}}%
          \includegraphics[width=\linewidth]{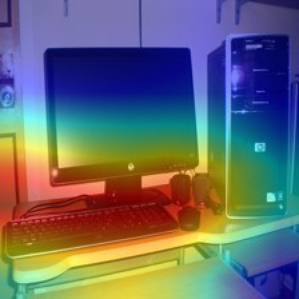}\\
          \vspace{0.1em}
          \caption*{Raw Image}
    \end{minipage}
    \hspace{-0.2em}
    \begin{minipage}[b]{0.12\textwidth} 
          \centering 
          \includegraphics[width=\linewidth]{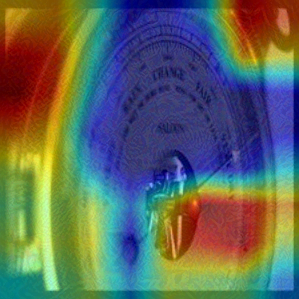}\\
          \vspace{0.1em}
          \includegraphics[width=\linewidth]{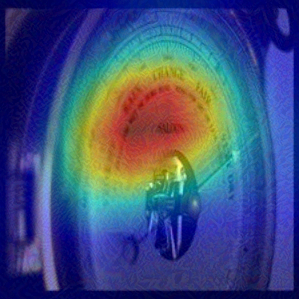}\\
          \vspace{0.1em}
          \includegraphics[width=\linewidth]{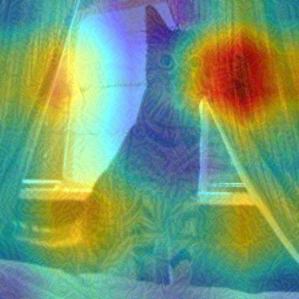}\\
          \vspace{0.1em}
          \includegraphics[width=\linewidth]{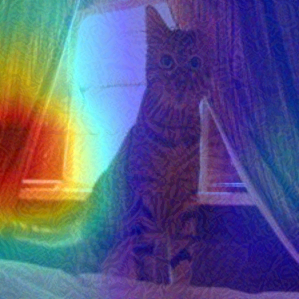}\\
          \vspace{0.1em}
          \includegraphics[width=\linewidth]{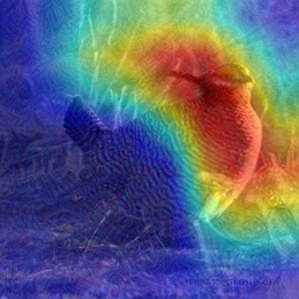}\\
          \vspace{0.1em}
          \includegraphics[width=\linewidth]{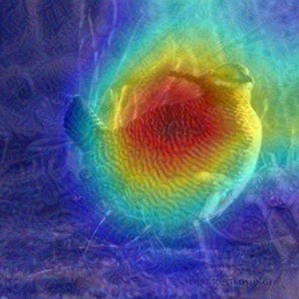}\\
          \vspace{0.1em}
          \includegraphics[width=\linewidth]{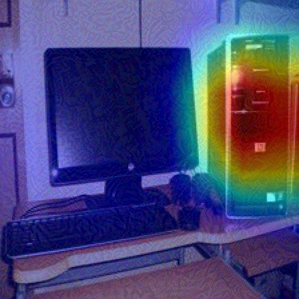}\\
          \vspace{0.1em}
          \includegraphics[width=\linewidth]{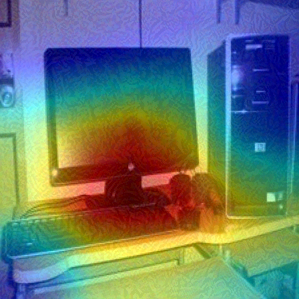}\\
          \vspace{0.1em}
          \caption*{DIM}
    \end{minipage}
    \hspace{-0.2em}
    \begin{minipage}[b]{0.12\textwidth} 
          \centering 
          \includegraphics[width=\linewidth]{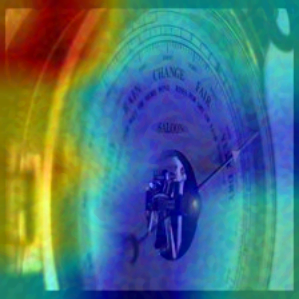}\\
          \vspace{0.1em}
          \includegraphics[width=\linewidth]{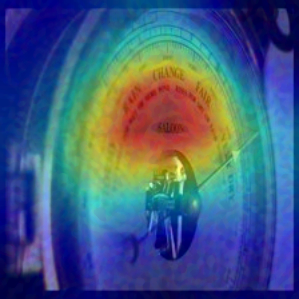}\\
          \vspace{0.1em}
          \includegraphics[width=\linewidth]{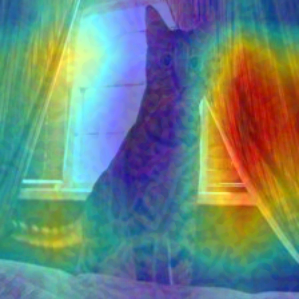}\\
          \vspace{0.1em}
          \includegraphics[width=\linewidth]{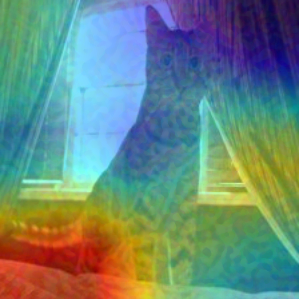}\\
          \vspace{0.1em}
          \includegraphics[width=\linewidth]{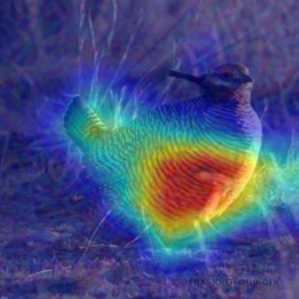}\\
          \vspace{0.1em}
          \includegraphics[width=\linewidth]{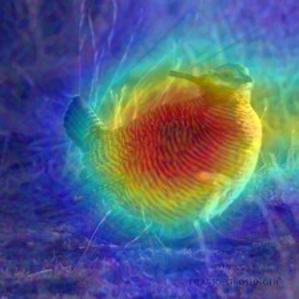}\\
          \vspace{0.1em}
          \includegraphics[width=\linewidth]{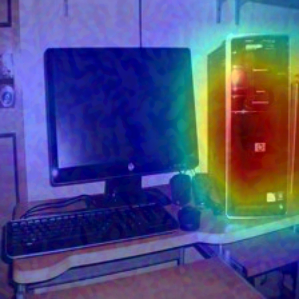}\\
          \vspace{0.1em}
          \includegraphics[width=\linewidth]{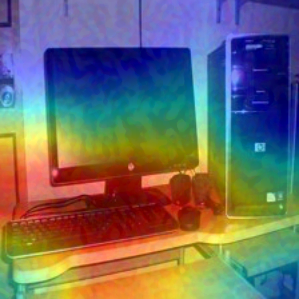}\\
          \vspace{0.1em}
          \caption*{TIM}
    \end{minipage}
     \hspace{-0.2em}
    \begin{minipage}[b]{0.12\textwidth} 
          \centering 
          \includegraphics[width=\linewidth]{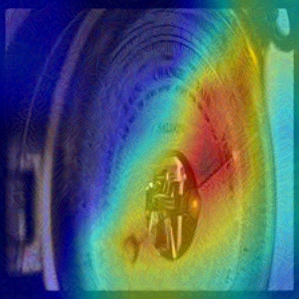}\\
          \vspace{0.1em}
          \includegraphics[width=\linewidth]{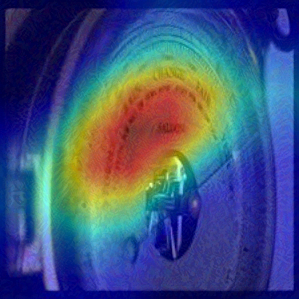}\\
          \vspace{0.1em}
          \includegraphics[width=\linewidth]{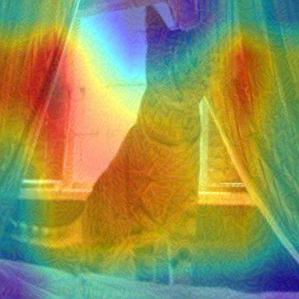}\\
          \vspace{0.1em}
          \includegraphics[width=\linewidth]{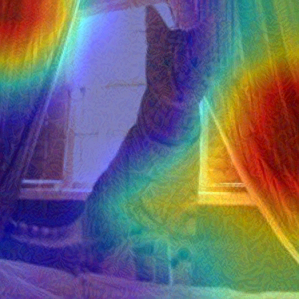}\\
          \vspace{0.1em}
          \includegraphics[width=\linewidth]{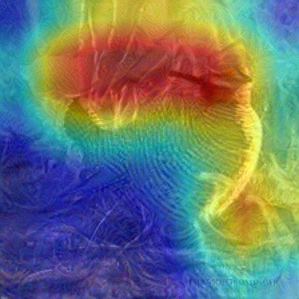}\\
          \vspace{0.1em}
          \includegraphics[width=\linewidth]{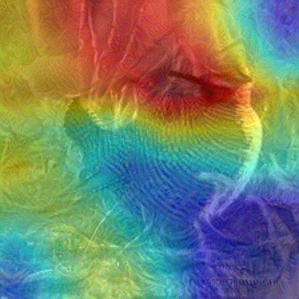}\\
          \vspace{0.1em}
          \includegraphics[width=\linewidth]{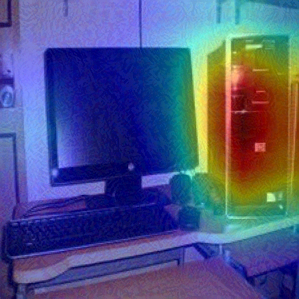}\\
          \vspace{0.1em}
          \includegraphics[width=\linewidth]{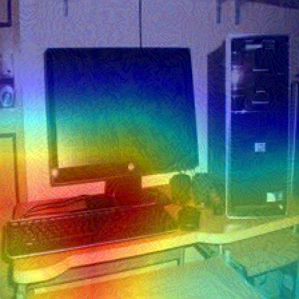}\\
          \vspace{0.1em}
          \caption*{SIM}
    \end{minipage}
    \hspace{-0.2em}
    \begin{minipage}[b]{0.12\textwidth} 
          \centering 
          \includegraphics[width=\linewidth]{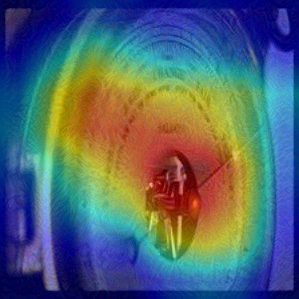}\\
          \vspace{0.1em}
          \includegraphics[width=\linewidth]{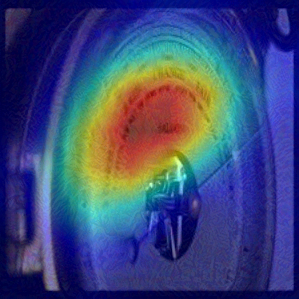}\\
          \vspace{0.1em}
          \includegraphics[width=\linewidth]{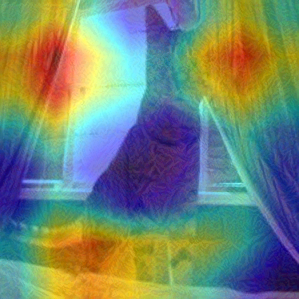}\\
          \vspace{0.1em}
          \includegraphics[width=\linewidth]{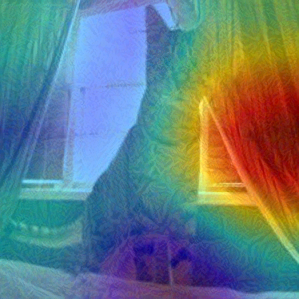}\\
          \vspace{0.1em}
          \includegraphics[width=\linewidth]{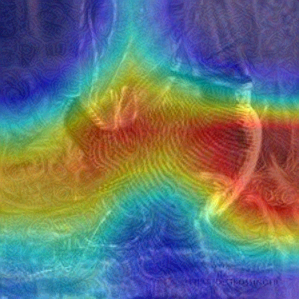}\\
          \vspace{0.1em}
          \includegraphics[width=\linewidth]{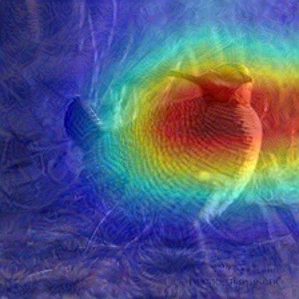}\\
          \vspace{0.1em}
          \includegraphics[width=\linewidth]{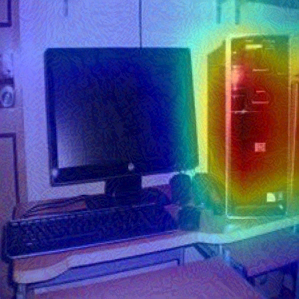}\\
          \vspace{0.1em}
          \includegraphics[width=\linewidth]{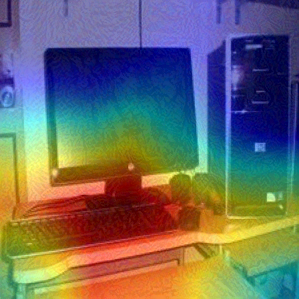}\\
          \vspace{0.1em}
          \caption*{\textit{Admix}}
    \end{minipage}
    \hspace{-0.2em}
    \begin{minipage}[b]{0.12\textwidth} 
          \centering 
          \includegraphics[width=\linewidth]{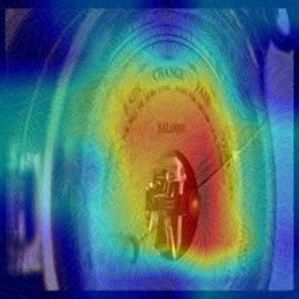}\\
          \vspace{0.1em}
          \includegraphics[width=\linewidth]{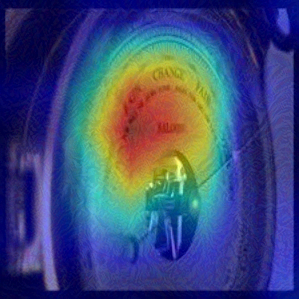}\\
          \vspace{0.1em}
          \includegraphics[width=\linewidth]{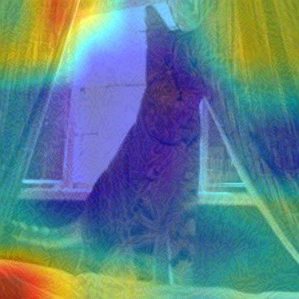}\\
          \vspace{0.1em}
          \includegraphics[width=\linewidth]{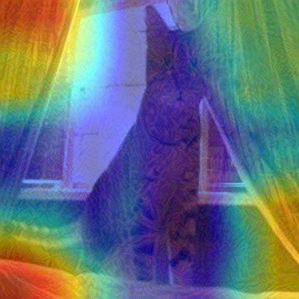}\\
          \vspace{0.1em}
          \includegraphics[width=\linewidth]{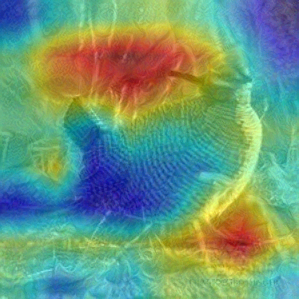}\\
          \vspace{0.1em}
          \includegraphics[width=\linewidth]{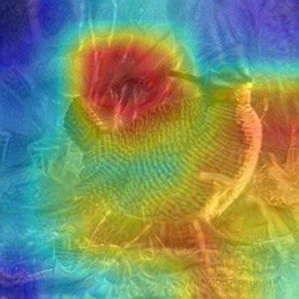}\\
          \vspace{0.1em}
          \includegraphics[width=\linewidth]{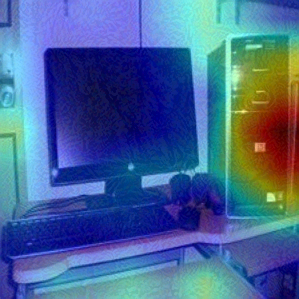}\\
          \vspace{0.1em}
          \includegraphics[width=\linewidth]{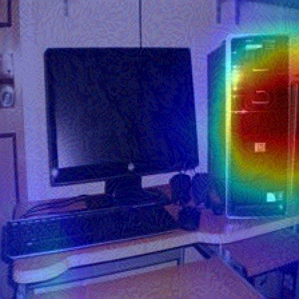}\\
          \vspace{0.1em}
          \caption*{PAM}
    \end{minipage}
    \hspace{-0.2em}
    \begin{minipage}[b]{0.12\textwidth} 
          \centering 
          \includegraphics[width=\linewidth]{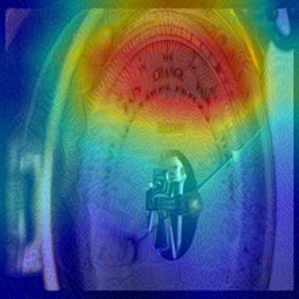}\\
          \vspace{0.1em}
          \includegraphics[width=\linewidth]{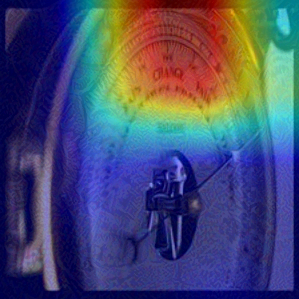}\\
          \vspace{0.1em}
          \includegraphics[width=\linewidth]{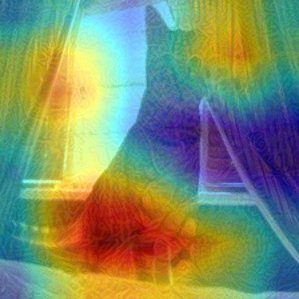}\\
          \vspace{0.1em}
          \includegraphics[width=\linewidth]{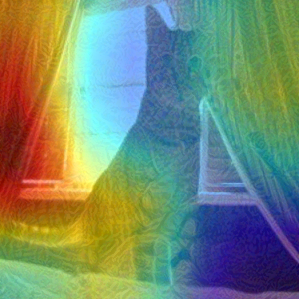}\\
          \vspace{0.1em}
          \includegraphics[width=\linewidth]{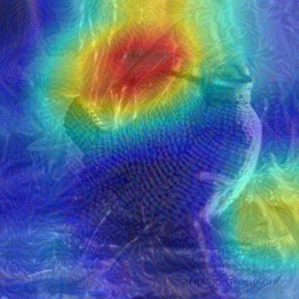}\\
          \vspace{0.1em}
          \includegraphics[width=\linewidth]{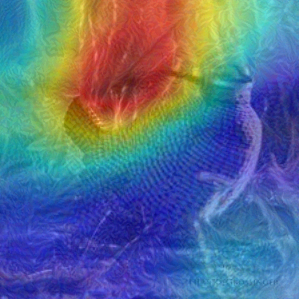}\\
          \vspace{0.1em}
          \includegraphics[width=\linewidth]{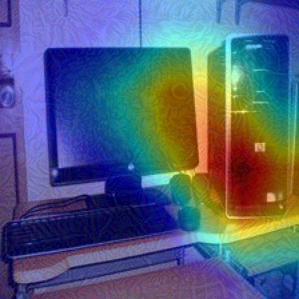}\\
          \vspace{0.1em}
          \includegraphics[width=\linewidth]{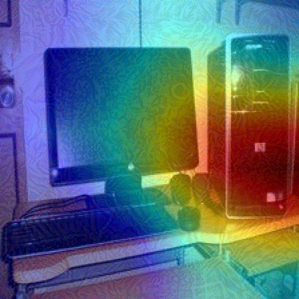}\\
          \vspace{0.1em}
          \caption*{\name}
    \end{minipage}
    \caption{The visualization of attention heatmaps~\cite{Selvaraju2017gradcam} of adversaries crafted by various input transformation methods on source model Inc-v3~\cite{szegedy2017inceptionv4}, and target model Inc-v4~\cite{szegedy2016inceptionv3}}
    \label{fig:hm_app}
\end{figure*}

To further support our motivation, we visualize more attention heatmaps of sampled benign images on both source model Inc-v4 and target model Inc-v3 in Fig.~\ref{fig:hm_app}. We can observe that the attention heatmaps of adversarial examples crafted by \name on source model are more similar with heatmap on target model than the results of other attacks. These results validate our motivation that our proposed method can eliminate the variance of attention heatmaps among various models, which is of benefit to generate more transferable adversarial examples.

\newpage
\begin{table*}[tb]
\begin{center}
\begin{tabular}{cccccccc}
\toprule
Attack  & Inc-v3 & IncRes-v2 & Res-101 & Inc-v3$_{ens3}$ & Inc-v3$_{ens4}$ & IncRes-v2$_{ens}$\\
\midrule
BSR-DIM  & 96.8\imp{24.8} & 94.1\imp{30.3} & 87.8\imp{30.4} & 59.7\imp{37.1} & 53.6\imp{32.5} & 37.6\imp{11.7} \\
BSR-TIM & 93.8\imp{34.7} & 91.7\imp{42.7} & 84.6\imp{42.7} & 69.4\imp{42.6} & 63.4\imp{40.5} & 51.1\imp{34.5} \\
BSR-SIM &99.0\imp{18.6}&97.9\imp{24.5}&95.2\imp{25.8}&86.7\imp{38.1}&85.2\imp{40.0}&70.2\imp{40.6}\\
BSR-\textit{Admix}&99.1\imp{10.1}&98.5\imp{13.2}&97.0\imp{18.0}&88.0\imp{32.5}&86.0\imp{34.3}&74.0\imp{41.7}\\
BSR-PAM&99.2\imp{12.5 }&98.2\imp{16.6}  &96.2\imp{20.3}  &87.2\imp{31.8}  &81.4\imp{30.9}  &62.8\imp{29.6}\\
\midrule
\textit{Admix}-TI-DIM & 91.0 & 88.6 & 83.2 & 76.0 & 74.7 & 64.3\\
PAM-TI-DIM & 91.5 & 88.5 & 83.5 & 77.1 & 73.2 & 62.1\\
BSR-TI-DIM&91.2&88.5&83.5&77.1&73.2&62.1\\
BSR-SI-TI-DIM&\textbf{98.1}&\textbf{96.2}&\textbf{94.7}&\textbf{90.1}&\textbf{89.0}&\textbf{80.0}\\
\bottomrule
\end{tabular}
\vspace{-0.3em}
\caption{Attack success rates (\%) on seven models under single model setting with various input transformations combined with \name. The adversaries are crafted on Inc-v4. \textcolor{Red}{$\uparrow$} indicates the increase of attack success rate when combined with \name.}
\label{tab:combined_incv4}
\end{center}
\vspace{-1.5em}
\end{table*}
\begin{table*}[tb]
\begin{center}
\begin{tabular}{cccccccc}
\toprule
Attack  & Inc-v3 & Inc-v4 & Res-101 & Inc-v3$_{ens3}$ & Inc-v3$_{ens4}$ & IncRes-v2$_{ens}$\\
\midrule
BSR-DIM  & 95.1\imp{24.8} & 95.0\imp{30.3} & 91.5\imp{33.5} & 71.7\imp{41.3} & 65.4\imp{41.9} & 54.8\imp{37.9} \\
BSR-TIM & 93.4\imp{31.2} & 93.2\imp{37.6} & 89.1\imp{38.8} & 80.3\imp{47.9} & 75.9\imp{48.4} & 70.0\imp{47.4} \\
BSR-SIM & 98.5\imp{12.6}&98.4\imp{18.4}&97.4\imp{21.3}&92.5\imp{36.3}&89.5\imp{40.4}&81.9\imp{39.4}\\
BSR-\textit{Admix}&99.1\imp{~8.3 }&99.3\imp{13.0}&98.1\imp{15.9}&93.7\imp{30.1}&92.1\imp{35.5}&87.3\imp{37.9}\\
BSR-PAM &99.3\imp{10.7}  &99.2\imp{12.9}  &98.5\imp{16.9}  &94.2\imp{28.2}  &91.2\imp{32.9}  &83.3\imp{32.3}\\
\midrule
\textit{Admix}-TI-DIM & 90.9 & 89.5 & 86.5 & 81.7 & 77.7 & 76.4\\
PAM-TI-DIM&92.2&90.5&86.9&84.2&80.9&78.4\\
BSR-TI-DIM&94.5&93.6&90.6&81.0&77.5&72.9\\
BSR-SI-TI-DIM&\textbf{98.6}&\textbf{98.1}&\textbf{96.3}&\textbf{95.4}&\textbf{94.0}&\textbf{91.7}\\
\bottomrule
\end{tabular}
\vspace{-0.3em}
\caption{Attack success rates (\%) on seven models under single model setting with various input transformations combined with \name. The adversaries are crafted on IncRes-v2. \textcolor{Red}{$\uparrow$} indicates the increase of attack success rate when combined with \name.}
\label{tab:combined_incres}
\end{center}
\vspace{-1.5em}
\end{table*}
\begin{table*}[tb]
\begin{center}
\begin{tabular}{cccccccc}
\toprule
Attack  & Inc-v3 & Inc-v4 & IncRes-v2 & Inc-v3$_{ens3}$ & Inc-v3$_{ens4}$ & IncRes-v2$_{ens}$\\
\midrule
BSR-DIM  & 97.7\imp{21.7} & 96.6\imp{28.2} & 97.1\imp{26.8} & 82.7\imp{48.0} & 75.8\imp{44.0} & 59.7\imp{40.1} \\
BSR-TIM & 96.7\imp{36.8} & 95.6\imp{43.4} & 94.9\imp{43.0} & 86.6\imp{52.2} & 83.4\imp{52.2} & 75.3\imp{51.6} \\
BSR-SIM & 98.8 \imp{20.3}&98.3\imp{25.8}&98.1\imp{26.9}&91.5\imp{46.7}&89.1\imp{49.0}&75.8\imp{48.5}\\
BSR-\textit{Admix} & 99.6\imp{15.1}&99.2\imp{19.0}&99.4\imp{18.7}&94.2\imp{42.6}&92.5\imp{47.8}&82.4\imp{52.5}\\
BSR-PAM &98.7\imp{21.3 }  &97.2\imp{23.3}  &97.7\imp{ 22.0}  &90.0\imp{38.8}  &86.8\imp{40.3}  &70.3\imp{38.1}\\
\midrule
\textit{Admix}-TI-DIM & 89.5 & 85.6 & 87.5 & 80.4 & 75.2& 67.5\\
PAM-TI-DIM&85.8&83.2&84.9&81.1&76.1&66.9\\
BSR-TI-DIM&97.3&95.1&95.6&88.3&85.4&76.1\\
BSR-SI-TI-DIM&\textbf{97.4}&\textbf{97.0}&\textbf{96.6}&\textbf{94.8}&\textbf{93.2}&\textbf{89.1}\\
\bottomrule
\end{tabular}
\vspace{-0.3em}
\caption{Attack success rates (\%) on seven models under single model setting with various input transformations combined with \name. The adversaries are crafted on Res-101. \textcolor{Red}{$\uparrow$} indicates the increase of attack success rate when combined with \name.}
\label{tab:combined_res}
\end{center}
\vspace{-1.5em}
\end{table*}


\end{document}